\documentclass{article}

\usepackage{amsmath,bm}
\usepackage{amsfonts}
\usepackage{color}
\usepackage{graphicx}
\usepackage{subcaption}
\usepackage[final]{corl_2017} 

\def\AA{\mathcal{A}}
\def\DD{\mathcal{D}}
\def\HH{\mathcal{H}}

\def\MM{\mathcal{M}}

\def\SS{\mathcal{S}}

\newcommand{\dr}{\mathrm{\,d}}

\newcommand{\reals}{{\bf R}}

\newcommand{\symm}{{\mbox{\bf S}}}  
\newcommand{\pdot}{\mathord{\cdot}} 

\newcommand{\E}{\mathop{\bf E{}}}

\newcommand{\argmin}{\mathop{\rm argmin}}

\newcommand{\minimize}{\mathop{\rm minimize}}
\newcommand{\maximize}{\mathop{\rm maximize}}

\newcommand{\cf}{{\it cf.}}
\newcommand{\eg}{{\it e.g.}}
\newcommand{\ie}{{\it i.e.}}

\title{Manifold Regularization for Kernelized LSTD }

\author{
  Xinyan Yan \\
  \texttt{xinyanyan@google.com}
  \And
  Krzysztof Choromanski \\
  \texttt{kchoro@google.com} \\ \\
  \And
  Byron Boots \\
  \texttt{bboots@google.com } \\
  \And
  Vikas Sindhwani \\
  \texttt{sindhwani@google.com}
  }

\begin{document}
\maketitle

\begin{abstract}
  Policy evaluation or value function or Q-function approximation is a key procedure in reinforcement learning (RL). It is a necessary component of policy iteration and can be used for variance reduction in policy gradient methods. Therefore its quality has a significant impact on most RL algorithms.
  Motivated by manifold regularized learning, we propose a novel kernelized policy evaluation method that takes advantage of the intrinsic geometry of the state space learned from data, in order to achieve better sample efficiency and higher accuracy in Q-function approximation.
 Applying the proposed method in the Least-Squares Policy Iteration (LSPI) framework, we observe superior performance compared to widely used parametric basis functions on two standard benchmarks in terms of policy quality.
 
\end{abstract}


\section{Problem description}
We consider discrete-time infinite horizon discounted Markov Decision Process (MDP),
with states $s \in \SS$, admissible actions $a \in \AA$, discount factor $\gamma \in (0,1)$, reward function $r:\SS \times \AA \rightarrow \reals$,
and transition probability density $p(s'|s,a)$.
Solving the MDP means finding a policy that maximizes the accumulated discounted rewards with states $s$ and actions $a$ distribution induced by applying that policy:
\begin{gather} \label{eq:rl_problem}
  \begin{array}{ll}
    \maximize_\pi & \E_{s,a \sim p_\pi} \sum_{i=0}^{\infty} \gamma^i r(s_i,a_i).
  \end{array}  
\end{gather}
Policy evaluation is a key component of policy iteration and policy gradient approaches to~\eqref{eq:rl_problem}, with the
objective of finding the Q-function\footnote{Throughout this work, we consider Q-function, instead of state value function, in order to cover the model-free setting.} $Q^\pi: \SS \times \AA \rightarrow \reals$ associated with a stationary deterministic policy $\pi: \SS \rightarrow \AA $, which is the fixed point of the Bellman operator $T^\pi$ or the solution of the Bellman equation:
\begin{gather} \label{eq:fixed_point}
  Q(s,a) = T^\pi Q (s,a) := r(s,a) + \E_{s'|s,a} \gamma Q(s', \pi(s')), \quad \forall s, a.
\end{gather}
When the state space is large or infinite, solving \eqref{eq:fixed_point} exactly becomes intractable.
Least-Squares Temporal Difference (LSTD) is a widely used simulation-based algorithm for 
approximately solving the projected Bellman equation 
with linear representation of value function~\citep{bradtke1996linear}.
In order to apply LSTD to policy iteration, \citet{lagoudakis2003least} proposed a Q-function extension,
and showed that the resulting policy iteration algorithm can be successfully applied to control problems.

In this work, we study LSTD-based approximate policy evaluation methods that benefit from manifold regularized learning, with the intuition that the Q-function is smooth on the manifold where states lie and not necessary in the ambient Euclidean space.
Such manifold structure naturally arises in many robotics tasks due to constraints in the state space.
For example, contact, \eg, between foot and the ground or a hand and an object~\citep{posa2016optimization}, restricts feasible states to lie along a manifold or a union of manifolds.
Other examples include the cases when state variables belong to Special Euclidean group SE(3) to encode 3D poses, and obstacle avoidance settings where geodesic paths between state vectors are naturally better motivated than geometrically infeasible straight-line paths in the ambient space.

We provide a brief introduction to manifold regularized learning and a kernelized LSTD approach in \S\ref{sec:background}. Our proposed method is detailed in \S\ref{sec:approach} and experimental results are presented in \S\ref{sec:experiment}. Finally \S\ref{sec:related_work} and \S\ref{sec:conclusion} conclude the paper with related and future work.

\section{Background} \label{sec:background}

\subsection{Manifold regularized learning} \label{sec:manifold_reg}
Manifold regularization has previously been studied in semi-supervised learning~\citep{belkin2006manifold}.
This data-dependent regularization exploits the geometry of the input space, thus achieving better generalization error. 
Given a labeled dataset $\DD=\{(x_i, y_i)\}_{i=1}^n$, Laplacian Regularized Least-Squares method (LapRLS)~\cite{belkin2006manifold}, finds a function $f$ in a reproducing kernel Hilbert space $\HH$ that fits the data with proper complexity for generalization: 
\begin{gather} \label{eq:manifold_obj}
  \begin{array}{ll}
    \minimize_{f\in \HH} & \frac{1}{n} \|Y-f(X)\|_2^2+
                      \lambda_f \|f\|_{\HH}^2 + \frac{1}{n^2}\lambda_{\MM} \|f\|_{\MM}^2,
  \end{array}
\end{gather}
where data matrices $X = \begin{bmatrix} x_1 & \ldots & x_n \end{bmatrix}^T$ and $Y = \begin{bmatrix} y_1 & \ldots & y_n \end{bmatrix}^T$
with $x_i \in \reals^d$ and $y_i \in \reals$,
$\|\pdot\|_\HH$ is the norm in $\HH$ and $\|\pdot\|_{\MM}$ is a penalty that reflects the intrinsic structure of $P_X$ (see \S{2.2} in~\citep{belkin2006manifold} for choices of $\|\pdot\|_{\MM}$),
and scalars $\lambda_f$, $\lambda_{\MM}$ are regularization parameters\footnote{$P_X$ denotes the marginal probability distribution of inputs, and $f(X) = \begin{bmatrix}f(x_1) \ldots f(x_n) \end{bmatrix}^T$.}.
A natural choice of $\|\pdot\|_{\MM}$ is
$\|f\|_{\MM}^2 = \int_{x\in \MM} \|\nabla_{\MM} f\|^2 \dr P_X(x)$,
where $\MM$ is the support of $P_X$ which is assumed to be a compact manifold~\citep{belkin2006manifold}
and $\nabla_{\MM}$ is the gradient along $\MM$.
When $P_X$ is unknown as in most learning settings, $\|\pdot\|_{\MM}$ can be estimated empirically, 
and the optimization problem~\eqref{eq:manifold_obj} becomes
\begin{gather} \label{eq:manifold_finite}
  \begin{array}{ll}
    \minimize_{f\in\HH} & \frac{1}{n} \|Y-f(X)\|_2^2 + 
                      \lambda_f \|f\|_{\HH}^2 + \frac{1}{n^2}\lambda_{\MM} \|f(X)\|^2_L,
  \end{array}
\end{gather}
where $\|x\|_L^2 = x^T L x$ and matrix
$L \in \symm^{n}_+$
is a graph Laplacian\footnote{$\symm^{n}_+$ denotes the set of symmetric $n \times n$ positive semidefinite matrices.}
(different ways of constructing graph Laplacian from data can be found in ~\citep{belkin2003laplacian}). Note that problem \eqref{eq:manifold_finite} is still convex, 
since graph Laplacian $L$ is positive semidefinite, and the multiplicity of eigenvalue $0$ is the number of connected components in the graph.
In fact, a broad family of graph regularizers can be used in this context~\citep{smola2003kernels,zhou2011semi}. This includes the iterated graph Laplacian which is theoretically better than the standard Laplacian~\citep{zhou2011semi}, or the diffusion kernel on graphs.

Based on Representer Theorem~\citep{scholkopf2001generalized}, the solution of \eqref{eq:manifold_finite} has the form $f^\star(x) = \alpha^T k(X, x)$, where $\alpha \in \reals^n$, and $k$ is the kernel associated with $\HH$.
After substituting the form of $f^\star$ in \eqref{eq:manifold_finite}, and solving the resulting least-squares problem, we can derive
\begin{gather}
  \alpha = (K + \lambda_f nI + \lambda_{\MM} \frac{1}{n} L K)^{-1} Y,
\end{gather}
where matrix $K \in \symm^n_+$ is the gram matrix, and matrix $I$ is an identity matrix of matching size.

\subsection{Kernelized LSTD with $\ell_2$ regularization} \label{sec:reg-lstd}

\citet{farahmand2009regularized} introduce
an $\ell_2$ regularization extension to kernelized LSTD~\citep{xu2007kernel}, termed Regularized LSTD (REG-LSTD),
featuring better control of the complexity of the function approximator through regularization, and mitigating the burden of selecting basis functions through kernel machinery.
REG-LSTD is formulated by adding $\ell_2$ regularization terms to both steps in a nested minimization problem that is equivalent to the original LSTD~\citep{antos2008learning}:
\begin{align}
  h_Q &= \argmin_{h \in \HH} \E_{s,a}  \left(h(s,a) - T^\pi Q(s,a)\right)^2
  + \lambda_h \|h\|^2_{\HH}, \label{eq:h_Q}\\  
  Q^\star &=  \argmin_{Q \in \HH}  \E_{s,a} \left( Q(s,a) - h_Q(s,a) \right)^2 + \lambda_Q \|Q\|^2_{\HH}, \label{eq:Q_pi}
\end{align}
where informally speaking, $h_Q$ in \eqref{eq:h_Q} is the  projection of $T^\pi Q$ \eqref{eq:fixed_point} on $\HH$,
and $Q$ is enforced to be close to that projection in \eqref{eq:Q_pi}. 
The kernel $k$ can be selected as a tensor product kernel $k(x,x') = k_1(s,s)k_2(a,a')$, where $x = (s, a)$ (note that the multiplication of two kernels is a kernel~\citep{genton2001classes}).
Furthermore, $k_2$ can be the Kronecker delta function, if the admissible action set $\AA$ is finite. 

As in LSTD, the expectations in \eqref{eq:h_Q} and \eqref{eq:Q_pi} are then approximated with finite samples $\DD = \{(s_i, a_i, s'_i, r_i)\}_{i=1}^n$,
leading to
\begin{align}
  {h}_Q &= \argmin_{h \in \HH} \frac{1}{n} \|h(X) - R -\gamma Q(X')\|^2_2
  + \lambda_h \|h\|^2_{\HH}  \label{eq:h_Q_finite} \\
  Q^\star &=  \argmin_{Q \in \HH}  \frac{1}{n} \| Q(X) - h_Q(X) \|_2^2 + \lambda_Q \|Q\|^2_{\HH}, \label{eq:Q_pi_finite}
\end{align}
where $X$ is defined as in \eqref{eq:manifold_obj} with $x_i = (s_i, a_i)$, similarly $X'$ consists of $x'_i = (s'_i, \pi(s'_i))$ with actions generated by the policy $\pi$ to evaluate,
and reward vector $R = r(X)$.
Invoking Representer theorem~\citep{scholkopf2001generalized}, \citet{farahmand2009regularized} showed that
\begin{gather} \label{eq:rep_reg_lstd}
  h_Q(x) = \beta^T k(X, x), \quad Q^{\star}(X) = \alpha^T k(\tilde X, x)
\end{gather}
where weight vectors $\beta \in \reals^n$ and $\alpha \in \reals^{2n}$, and matrix $\tilde X$ is $X$ and $X'$ stacked. Substituting
\eqref{eq:rep_reg_lstd}  in \eqref{eq:h_Q} and \eqref{eq:Q_pi}, we obtain the formula for $\alpha$:
\begin{gather} \label{eq:opt_alpha}
  \alpha = (F^T F K_Q + \lambda_Q nI) ^{-1} F^T E R,
\end{gather}
where $F = C_1-\gamma E C_2$, $E = K_h(K_h + \lambda_h n I)^{-1}$, $K_Q = K(\tilde X, \tilde X)$, $K_h = K(X, X)$, 
$C_1 = \begin{bmatrix} I & 0 \end{bmatrix}$, and $C_2 = \begin{bmatrix} 0 & I \end{bmatrix}$.

\section{Our approach} \label{sec:approach}
Our approach combines the benefits of both manifold regularized learning (\S\ref{sec:manifold_reg}) that incorporates geometric structure, and kernelized LSTD with $\ell_2$ regularization (\S\ref{sec:reg-lstd}) that offers better control over function complexity and eases feature selection.

More concretely, besides the regularization term in \eqref{eq:Q_pi_finite} that controls the complexity of $Q$ in the ambient space, we augment the objective function with a manifold regularization term that enforces $Q$ to be smooth on the manifold that supports $P_X$.
In particular, if the manifold regularization term is chosen as in \S\ref{sec:manifold_reg},
large penalty is imposed if $Q$ varying too fast along the manifold. 
With the empirically estimated manifold regularization term, optimization problem \eqref{eq:Q_pi_finite} becomes (\cf, \eqref{eq:manifold_finite})
\begin{gather}
  Q^\star =  \argmin_{Q \in \HH}  \frac{1}{n} \| Q(X) - h_Q(X) \|_2^2 + \lambda_Q \|Q\|^2_{\HH}  +
  \lambda_{\MM} \frac{1}{(2n)^2} \|Q(\tilde X)\|^2_L,
\end{gather}
which admits the optimal weight vector $\alpha$ (\cf, \eqref{eq:opt_alpha}):
\begin{gather}
  \alpha = (F^T F K_Q + \lambda_Q nI + \frac{1}{4n} \lambda_{\MM} L K_Q) ^{-1} F^T E R.
\end{gather}

\section{Experimental results} \label{sec:experiment}
We present experimental results on two standard RL benchmarks: two-room
navigation and cart-pole balancing.
REG-LSTD with manifold regularization (MR-LSTD) (\S\ref{sec:approach})
is compared against REG-LSTD without manifold regularization (\S\ref{sec:reg-lstd}) and LSTD with three commonly used basis function construction mechanisms: polynomial~\citep{schweitzer1985generalized, lagoudakis2003least}, radial basis functions (RBF)~\citep{lagoudakis2003least,sugiyama2008geodesic}, and Laplacian eigenmap~\citep{mahadevan2012representation,petrik2007analysis}, in terms of the quality of the policy produced by Least-Squares Policy Iteration (LSPI)~\citep{lagoudakis2003least}. 
The kernel used in the experiments is $k(x,x') = \exp(-\frac{\|s-s'\|_2^2}{2\sigma^2}) \delta(a-a')$ with hyperparameter $\sigma$. 
We use combinatorial Laplacian with adjacency matrix computed from $\epsilon$-neighborhood with $\{0,1\}$ weights~\citep{belkin2003laplacian}.

\subsection{Two-room navigation}
The two-room navigation problem is a classic benchmark for RL algorithms that cope with either discrete~\citep{petrik2007analysis, parr2008analysis, mahadevan2012representation}
or continuous state space~\citep{sugiyama2008geodesic,taylor2009kernelized}.
In the vanilla two-room problem, the state space is a $10 \times 10$ discrete grid world, and admissible actions are stepping in one of the four cardinal directions, \ie, up, right, down, and left.
The dynamics is stochastic: each action succeeds with probability $0.9$ when movement is not blocked by obstacle or border, otherwise leaves the agent in the same location.
The goal is to navigate to the diagonally opposite corner in the other room, with a single-cell doorway connecting the two rooms. Reward is $1$ at the goal location otherwise $0$, and the discount factor is set to $0.9$.
Data are collected beforehand by uniformly sampling states and actions, and used throughout LSPI iterations. 
Seen from Table~\ref{table:two_room}, Laplacian eigenmap which exploits intrinsic state geometry outperforms parametric basis functions: polynomial and RBF. The MR-LSTD method we propose achieves the best performance.


\begin{table}\centering
  \begin{tabular}{c| c c c c c}
    \# of samples & polynomial & RBF & eigenmap & REG-LSTD & MR-LSTD \\ 
    \hline
    250 & 7.00 & 16.31 & 6.44 & 6.69 & \emph{4.40} \\
    500 & 5.46 & 11.7 & 3.9 & 2.43 & \emph{0.56}  \\
    1,000 & 4.55 & 2.4 & 1.56 & 0.24 & \emph{0.02}  \\
  \end{tabular}
  \caption{\emph{Two-room navigation.} Quality of policies computed from LSPI-based methods with different basis functions,
    measured by the number of states whose corresponding actions are NOT optimal (the smaller the better), with varying number of samples and averaged over 100 runs.
    The maximum number of iterations of LSPI is 50. 
    Best performance among LSPI iterations and coarsely tuned hyperparameters is reported.
    In particular, polynomial degree ranges from $1$ to $8$, and RBF discretization per-dimension ranges from $2$ to $7$. 
  }
  \label{table:two_room}  
\end{table}

\subsection{Cart-pole Balancing}
The cart-pole balancing task is to balance a pole upright by applying force to the cart to which it's attached\footnote{The cart-pole environment in OpenAI Gym package is used in our implementation.}~\citep{barto1983neuronlike}.
The agent constantly receives $0$ reward until trial ends with $-1$ reward when it's $12^\circ$ away from the upright posture.
On the contrary to the two-room navigation task, the state space is continuous, consisting of angle and angular velocity of the pole. Admissible actions are finite: pushing the cart to left or right. 
The discount factor $\gamma = 0.99$.
Data are collected from random episodes, \ie, starting from a perturbed upright posture and applying uniformly sampled actions. 
Results are reported in Table~\ref{table:cartpole}, which shows that REG-LSTD achieves significantly better performance than parametric basis functions, and performance is even improved further with manifold regularization.

\begin{table}\centering
  \begin{tabular}{c| c c c c}
    \# of samples & polynomial & RBF & REG-LSTD & MR-LSTD \\ 
    \hline
    250 & 121.10 & 116.59 & 186.82 & \emph{195.78} \\
    500 & 150.50 & 112.22 & 181.44  & \emph{195.91}  \\
    1,000 & 154.38 & 130.74 & 188.82 & \emph{198.12}  \\
  \end{tabular}
  \caption{\emph{Cart-pole.} Quality of policies computed from LSPI-based methods with different basis functions, 
    measured by the average number of steps before termination over 100 trials (the larger the better). The maximum length of trials is set to 200. All the numbers are the average over 100 runs, and the LSPI iteration number and hyperparameter tuning are the same as in two-room navigation.\vspace{-3mm}
  }
  \label{table:cartpole}  
\end{table}

\section{Related work} \label{sec:related_work}
The closest work to this paper was recently introduced by~\citet{li2017manifold}, which utilized manifold regularization by learning state representation through unsupervised learning, and then adopting the learned representation in policy iteration.
In contrast to this work, we naturally blend manifold regularization with policy evaluation with possibly provable performance guarantee (left for future work).
There is also work on constructing basis functions directly from geometry, \eg, Laplacian methods~\citep{mahadevan2012representation,petrik2007analysis}, and 
geodesic Gaussian kernels~\citep{sugiyama2008geodesic}.
Furthermore, different regularization mechanisms to LSTD have been proposed, including $\ell_1$ regularization for feature selection~\citep{kolter2009regularization}, and nested $\ell_2$ and $\ell_1$ penalization to avoid overfitting~\citep{hoffman2012regularized}. 

\section{Conclusion and future work} \label{sec:conclusion}
We propose manifold regularization for a kernelized LSTD approach in order to exploit the intrinsic geometry of the state space for better sample efficiency and Q-function approximation, and demonstrate superior performance on two standard RL benchmarks. 
Future work directions include 
1) accelerating by structured random matrices for kernel machinery~\citep{felix2016orthogonal,choromanski2017unreasonable}, and graph sketching for graph regularizer construction to scale up to large datasets and rich observations, \eg, images,
2) providing theoretical justification, and combining manifold regularization with deep neural nets~\citep{mnih2015human,silver2014deterministic} and other policy evaluation, \eg, \citep{dai2016learning,xu2007kernel} and policy iteration algorithms, 
3) learning with a data-dependent kernel that capturing the geometry (equivalent to the manifold regularized solution~\citep{sindhwani2005beyond}) that makes it easier to derive new algorithms, 
and 4) extension to continuous action spaces by constructing kernels such that policy improvement (optimize over actions) is tractable~\citep{genton2001classes}.



\bibliography{references}
\end{document}